\documentclass[letterpaper, 10 pt, conference]{ieeeconf}  % Comment this line out if you need a4paper

\IEEEoverridecommandlockouts                              % This command is only needed if 
\usepackage{graphicx,multirow}
\usepackage{amssymb}
\usepackage{amsmath}
\usepackage{threeparttable}
\usepackage{xifthen}
\usepackage{xparse}
\usepackage{subdepth}
\usepackage{leftidx}
\usepackage{bm}
\usepackage{etoolbox}
\usepackage{amsmath}
\usepackage{multirow, makecell}

\newcommand{\norm}[1]{\left\Vert#1\right\Vert}
\newcommand{\abs}[1]{\left\vert#1\right\vert}

\newcommand{\bbm}{\begin{bmatrix}}
\newcommand{\ebm}{\end{bmatrix}}

\DeclareMathAlphabet{\mybf}{OT1}{ptm}{b}{n} % letters
\newcommand{\mybs}[1]{{\bm{#1}}} % symbols
\DeclareMathAlphabet{\mybfi}{OML}{cmm}{b}{it}

\newcommand{\mbf}[1]{
\ifcat\noexpand#1\relax % check if the argument is a control sequence
\mybs{#1}% probably Greek
\else
\mybf{#1}% single character
\fi
}

\newcommand{\mbfbar}[1]{{\overline{\mbf{#1}}}}
\newcommand{\mbfhat}[1]{{\hat{\mbf{#1}}}}
\newcommand{\mbftilde}[1]{{\tilde{\mbf{#1}}}}

\newcommand{\mbfdot}[1]{{\dot {\mbf{#1}}}}

\NewDocumentCommand{\mbfidentity}{o}{\IfValueTF{#1}{\mbf{I}_{#1\hspace{\rightshift}}}{\mbf{I}}}
\NewDocumentCommand{\mbfzero}{oo}{\IfValueTF{#1}{\mbf{0}_{#1\times#2\hspace{\rightshift}}}{\mbf{0}}}

\newcommand{\cframe}[1]{{\smash{\protect\underrightarrow{\mathcal{F}}_{#1}}}}

\newcommand{\homo}[1]{{\mybfi{#1}}}

\newcommand{\mbfh}[1]{{\homo{#1}}}

\newlength{\leftshift}
\newlength{\rightshift}
\newcommand{\pos}[2]{\leftidx{_{#1}}{ \mbf r}{_{#2\hspace{\rightshift}}}} % position
\newcommand{\posh}[2]{\leftidx{_{#1}}{\mbfh r}{_{#2\hspace{\rightshift}}}} % position in homogeneous representation
\NewDocumentCommand{\vel}{moo}{
	\IfValueTF{#1}{\leftidx{_{#1}}}{}{\mbf v}{\IfValueTF{#2}{_{#2#3\hspace{\rightshift}}}{}}}
\NewDocumentCommand{\veltilde}{moo}{
	\IfValueTF{#1}{\leftidx{_{#1}}}{}{\mbftilde v}{\IfValueTF{#2}{_{#2#3\hspace{\rightshift}}}{}}}
\NewDocumentCommand{\velbar}{moo}{
	\IfValueTF{#1}{\leftidx{_{#1}}}{}{\mbfbar v}{\IfValueTF{#2}{_{#2#3\hspace{\rightshift}}}{}}}
\NewDocumentCommand{\velhat}{moo}{
	\IfValueTF{#1}{\leftidx{_{#1}}}{}{\mbfhat v}{\IfValueTF{#2}{_{#2#3\hspace{\rightshift}}}{}}}
\NewDocumentCommand{\veldot}{moo}{
	\IfValueTF{#1}{\leftidx{_{#1}}}{}{\mbfdot v}{\IfValueTF{#2}{_{#2#3\hspace{\rightshift}}}{}}}

\NewDocumentCommand{\acc}{moo}{
	\IfValueTF{#1}{\leftidx{_{#1}}}{}{\mbf a}{\IfValueTF{#2}{_{#2#3\hspace{\rightshift}}}{}}}
\NewDocumentCommand{\acctilde}{moo}{
	\IfValueTF{#1}{\leftidx{_{#1}}}{}{\mbftilde a}{\IfValueTF{#2}{_{#2#3\hspace{\rightshift}}}{}}}
\NewDocumentCommand{\accbar}{moo}{
	\IfValueTF{#1}{\leftidx{_{#1}}}{}{\mbfbar a}{\IfValueTF{#2}{_{#2#3\hspace{\rightshift}}}{}}}
\NewDocumentCommand{\acchat}{moo}{
	\IfValueTF{#1}{\leftidx{_{#1}}}{}{\mbfhat a}{\IfValueTF{#2}{_{#2#3\hspace{\rightshift}}}{}}}
\NewDocumentCommand{\accdot}{moo}{
	\IfValueTF{#1}{\leftidx{_{#1}}}{}{\mbfdot a}{\IfValueTF{#2}{_{#2#3\hspace{\rightshift}}}{}}}

\NewDocumentCommand{\rotvel}{moo}{
	\IfValueTF{#1}{\leftidx{_{#1}}}{}{\mbf $\omega$}{\IfValueTF{#2}{_{#2#3\hspace{\rightshift}}}{}}}
\NewDocumentCommand{\rotveltilde}{moo}{
	\IfValueTF{#1}{\leftidx{_{#1}}}{}{\mbftilde $\omega$}{\IfValueTF{#2}{_{#2#3\hspace{\rightshift}}}{}}}
\NewDocumentCommand{\rotvelbar}{moo}{
	\IfValueTF{#1}{\leftidx{_{#1}}}{}{\mbfbar $\omega$}{\IfValueTF{#2}{_{#2#3\hspace{\rightshift}}}{}}}
\NewDocumentCommand{\rotvelhat}{moo}{
	\IfValueTF{#1}{\leftidx{_{#1}}}{}{\mbfhat $\omega$}{\IfValueTF{#2}{_{#2#3\hspace{\rightshift}}}{}}}
\NewDocumentCommand{\rotveldot}{moo}{
	\IfValueTF{#1}{\leftidx{_{#1}}}{}{\mbfdot $\omega$}{\IfValueTF{#2}{_{#2#3\hspace{\rightshift}}}{}}}

\newcommand{\T}[2]{{\mbfh T}{_{#1#2\hspace{\rightshift}}}} % homogeneous transformation matrix
\newcommand{\q}[2]{{\mbf q}{_{#1#2\hspace{\rightshift}}}} % quaternion of rotation
\title{\LARGE \bf
Uncertainty-Aware Visual-Inertial SLAM with \\Volumetric Occupancy Mapping 
}

\author{Jaehyung Jung$^{1}$, Simon Boche$^{1}$, Sebasti\'{a}n Barbas Laina$^{1}$, and Stefan Leutenegger$^{1}$% <-this % stops a space
\thanks{This work was supported by the European Commission Horizon Europe project AUTOASSESS (101120732) and DIGIFOREST (101070405).}% <-this % stops a space
\thanks{$^{1}$Smart Robotics Lab, School of Computation, Information and Technology (CIT); as well as Munich Institute of Robotics and Machine Intelligence (MIRMI) and Munich Center of Machine Learning (MCML), Technical University of Munich, Germany. {\tt\small firstname.surname@tum.de}}%
}

\begin{document}

\maketitle
\thispagestyle{empty}
\pagestyle{empty}

%%%%%%%%%%%%%%%%%%%%%%%%%%%%%%%%%%%%%%%%%%%%%%%%%%%%%%%%%%%%%%%%%%%%%%%%%%%%%%%%
\begin{abstract}

We propose visual-inertial simultaneous localization and mapping that tightly couples sparse reprojection errors, inertial measurement unit pre-integrals, and relative pose factors with dense volumetric occupancy mapping. Hereby depth predictions from a deep neural network are fused in a fully probabilistic manner.  Specifically, our method is rigorously \textit{uncertainty-aware}: first, we use depth and uncertainty predictions from a deep network not only from the robot's stereo rig, but we further probabilistically fuse motion stereo that provides depth information across a range of baselines, therefore drastically increasing mapping accuracy. Next, predicted and fused depth uncertainty propagates not only into occupancy probabilities but also into alignment factors between generated dense submaps that enter the probabilistic nonlinear least squares estimator. This submap representation offers globally consistent geometry at scale. Our method is thoroughly evaluated in two benchmark datasets, resulting in localization and mapping accuracy that exceeds the state of the art, while simultaneously offering volumetric occupancy directly usable for downstream robotic planning and control in real-time.

\end{abstract}

%%%%%%%%%%%%%%%%%%%%%%%%%%%%%%%%%%%%%%%%%%%%%%%%%%%%%%%%%%%%%%%%%%%%%%%%%%%%%%%%

\section{INTRODUCTION}
State estimation is a fundamental building block for downstream tasks including robot perception, human-robot interaction, and robot exploration. Cameras and inertial measurement units (IMUs) have become a typical minimalist sensor suite to approach state estimation problems by virtue of their complementary characteristics and ability to observe a metric scale. Visual-inertial simultaneous localization and mapping (VI-SLAM) is considered to be well-established with the recent advancement such as OKVIS~2~\cite{leutenegger2022okvis2}, ORB-SLAM3~\cite{campos2021orb}, OpenVINS~\cite{geneva2020openvins}, and VINS-Fusion~\cite{qin2019general}, to name a few. Reprojection or photometric errors formulated from a set of sparse or (semi-)dense landmarks are standard residuals in the VI-SLAM problem. However, this map representation is not suitable for higher-level robot tasks where densely populated free/occupied spaces should be modeled explicitly.

Dense mapping incrementally integrates onboard range or depth measurements given a sensor pose to a global model to build a volumetric map, which can use occupancy~\cite{SE2} or truncated signed distance field (TSDF)~\cite{newcombe2011kinectfusion, voxblox} as the underlying map representation. Given that the observation model plays an important role in the reconstruction quality, one has to decide how much the measured range of depth contributes to each voxel based on the uncertainty. Ideally, the uncertainty should be statistically consistent. Moreover, depth prediction with valid uncertainty or confidence has been an active research topic in computer vision~\cite{kendall2017uncertainties, poggi2020uncertainty}. Depending on the camera configuration, a monocular depth network estimates depths in a single image, a stereo network predicts disparity between left and right images, and a multi-view stereo (MVS) network estimates depths in a reference image frame given multiple posed images. However, in contrast to the recent developments of uncertainty learning in deep neural networks, a quadratic model where the uncertainty increases with the squared depth has been widely used in dense mapping. This model, which effectively assumes constant disparity uncertainty, is a good approximation but is not valid in general. For instance, the belief in a highly uncertain pixel, as typically the case in less-textured areas, should be discounted for safe robot perception.

\begin{figure}
\centerline{\includegraphics[width=\linewidth]{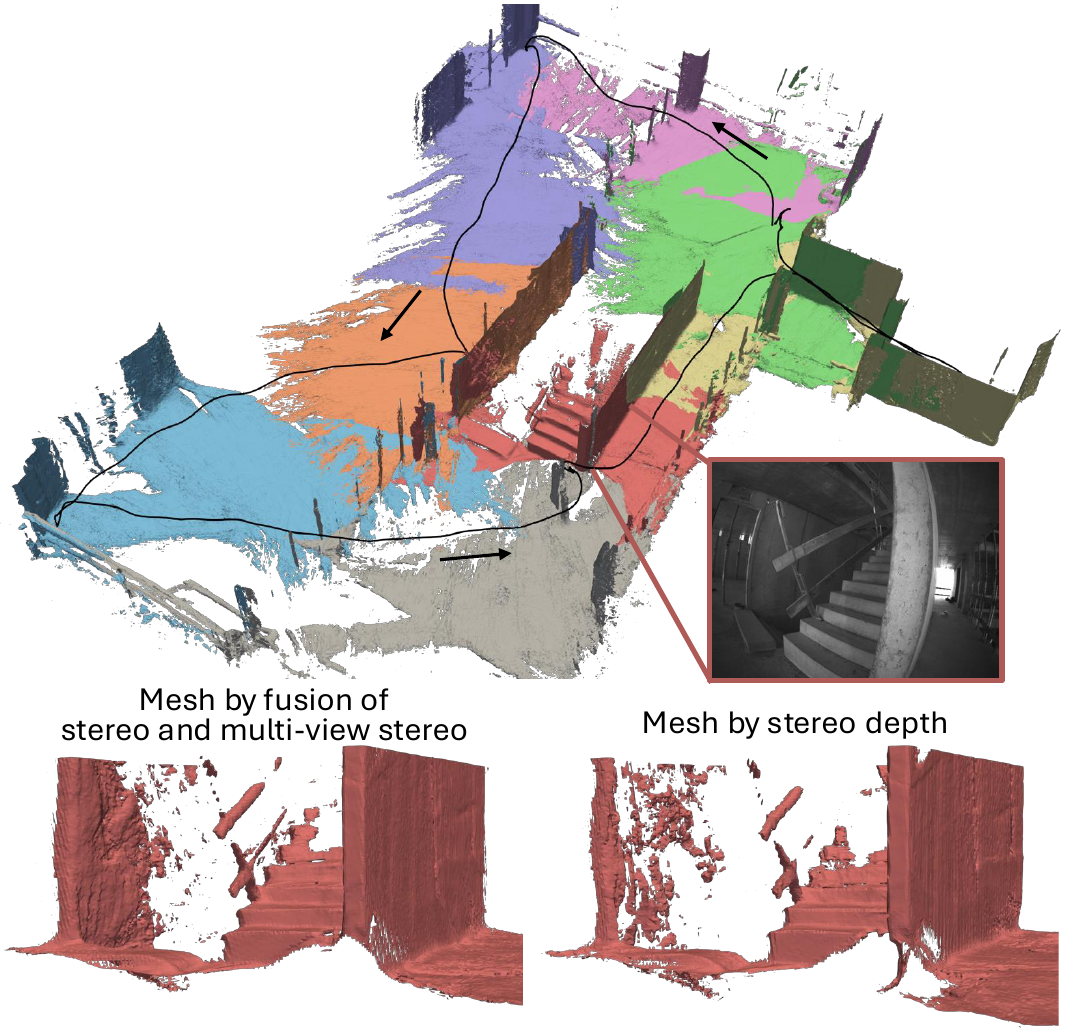}}
\caption{Our method estimates pose (black line) and volumetric occupancy represented in submaps (visualised here as colored meshes). We fuse stereo and MVS network depths based on their predicted uncertainties giving less noisy meshes, as can be qualitatively observed in the left mesh. Result from the Hilti-Oxford dataset~\cite{hilti2022}.} \label{fig:exp04_slam}
\end{figure}

To tackle this challenge, we propose VI-SLAM with volumetric occupancy mapping which is fully \textit{uncertainty-aware} in the sense that occupancy integration and the visual-inertial factor graph are properly weighted by the estimated uncertainty of \textit{depth fusion}. We optimally fuse depths from stereo and MVS networks based on their predicted uncertainties motivated by the complementary characteristics of small and large baselines from static and motion stereo. As shown in Fig. \ref{fig:exp04_slam}, our approach outputs real-time robot pose as well as volumetric occupancy maps that are being built incrementally where submaps are anchored at a local coordinate frame for globally consistent map representation. In summary, our contributions of this paper are as follows.
\begin{itemize}
    \item We introduce depth fusion of a deep MVS network and a stereo network based on their predicted network uncertainty to further improve the mapping accuracy.
    \item Leveraging the depth prediction and the associated uncertainty of a deep neural network, we formulate fully probabilistic VI-SLAM with volumetric occupancy submapping where the depth integration and the factor graph optimization are weighted by the predicted uncertainty.
    \item We thoroughly evaluate our approach in the benchmark datasets in terms of localization and mapping accuracy while achieving state-of-the-art performance running in real-time with a stereo-inertial setup.
\end{itemize}
\section{RELATED WORK}
We review the most relevant works in multi-view depth deep neural networks and dense SLAM including probabilistic mapping methods.

A stereo network aims to find the disparity in a rectified stereo image. Inspired by stereo matching~\cite{scharstein2002taxonomy}, the network typically follows three steps: features are extracted from the stereo pair, a cost volume is built along the horizontal direction, and then the disparity is computed with the following refinement or upsampling. ACVNet~\cite{xu2023accurate} proposed attention concatenation volume to emphasize the most useful values in the cost volume. Unimatch~\cite{xu2023unifying} presented a unified network for stereo matching, optical flow, and two-view depth prediction with feature enhancement by a transformer~\cite{dosovitskiy2020image}. Likewise, a deep MVS network follows a similar setup but with perspective warping based on camera geometry to construct the cost volume with unrectified images. MVSNet~\cite{yao2018mvsnet} aggregated multi-view features through a variance metric for the cost volume followed by 3D convolutions. Simplerecon~\cite{sayed2022simplerecon} circumvented 3D convolutions by incorporating camera geometry in the cost volume. Simplemapping~\cite{xin2023simplemapping} proposed to input a prior depth map from sparse landmarks to narrow down the depth plane. Leveraging the recent developments in deep neural networks for multi-view depth prediction, we tailor state-of-the-art methods~\cite{xu2023unifying, xin2023simplemapping} to learn the depth uncertainty via the Laplacian loss. The predicted uncertainty is utilized in our fully uncertainty-aware framework for depth fusion, occupancy mapping, and factor graph optimization.

Understanding dense geometry is indispensable for autonomous robots. KinectFusion~\cite{newcombe2011kinectfusion} pioneered this area with TSDF fusion and frame-to-model registration. ElasticFusion~\cite{whelan2016elasticfusion} models an environment with dense surfels, which are updated non-rigidly after a loop. CNN-SLAM~\cite{tateno2017cnn} and DeepFusion~\cite{laidlow2019deepfusion} employed a dense depth from a convolutional neural network and fuse depths from motion stereo to further improve the depth quality. Voxgraph~\cite{reijgwart2019voxgraph} advocates a submap representation for a scalable system and proposed submap alignment based on Euclidean distance fields  given odometry estimation. Regarding relevant works in an MVS network in SLAM, TANDEM~\cite{koestler2022tandem} integrated a deep MVS network in visual odometry, rendering additional depths from TSDF for image alignment. Simplemapping adopted Simplerecon~\cite{sayed2022simplerecon} and fused depths from the MVS network into a TSDF given poses and sparse landmarks from ORB-SLAM3~\cite{campos2021orb} but without any feedback from mapping. SigmaFusion~\cite{rosinol2023probabilistic} predicted dense depth uncertainty based on the information matrix from DroidSLAM~\cite{teed2021droid} and showed less noisy reconstruction through uncertainty integration. However, their method is memory intensive and the mapping is decoupled from the estimator as an ad-hoc mapping as opposed to ours. D3VO~\cite{yang2020d3vo} proposed to use the photometric uncertainty learned from depth estimation in direct visual odometry, but their method was not a full SLAM system. The most related work is~\cite{boche2024tightly} where an occupancy-to-point factor was introduced to align submaps. However, noisy depth measurements should be properly weighted for downstream tasks in contrast to precise LiDAR point clouds.

In contrast to previous works, our method tightly couples the visual-inertial estimator and volumetric occupancy with the occupancy-to-point factors for a globally consistent map. On top of this, our key contribution lies in the usage of uncertainty which is propagated to all downstream tasks.

\begin{figure}
\centerline{\includegraphics[width=\linewidth]{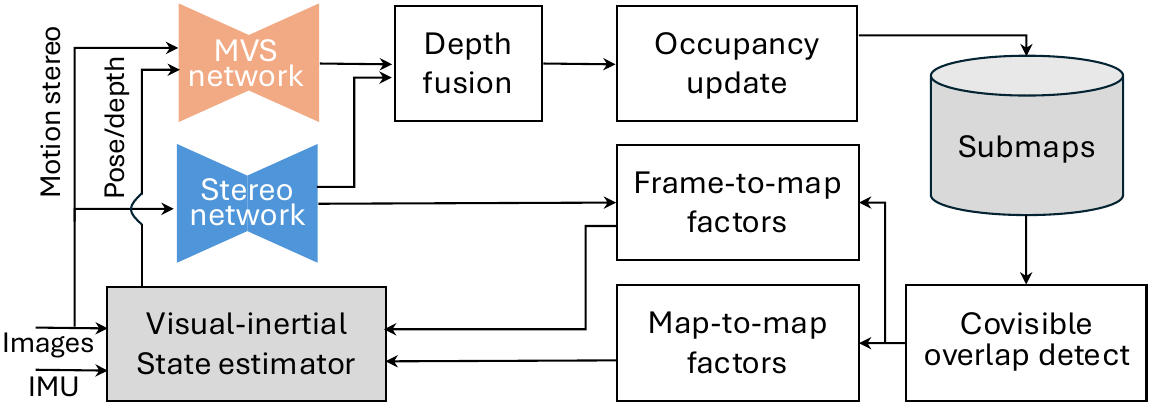}}
\caption{An overview block diagram of the proposed method. We fuse depths from stereo and MVS networks from which the current submap is expanded. Fused depth and stereo depth formulate map-to-map and frame-to-map factors in the visual-inertial estimator.} \label{fig:block}
\end{figure}
\section{DEFINITIONS AND NOTATIONS}
The objective of this work is to estimate robot poses as well as volumetric occupancy values represented in a set of submaps given inertial and multiple camera measurements based on their observation uncertainty. A \textit{submap} is defined as a local map representation with geometric overlaps to each other to reduce mapping error due to pose drift, since each submap's pose can be corrected after an update from the estimator~\cite{reijgwart2019voxgraph, laina2024scalable}. Our method is based on OKVIS2~\cite{leutenegger2022okvis2} for visual-inertial estimation and Supereight2~\cite{SE2} for occupancy submapping.

The world frame $\cframe{W}$ describes a gravity-aligned coordinate frame, the sensor frame $\cframe{S}$ is coincident with an IMU frame, and the $i$th camera frame $\cframe{C_i}$ is centered at the optical center of the corresponding camera. The $j$th submap frame $\cframe{M_j}$ is defined as the sensor frame when the new submap is declared. The robot state is represented by
\begin{equation}
    \mbf{x} = \begin{bmatrix}
        \pos{W}{S}^T & \q{W}{S}^T & \vel{W}^T & \mbf{b}_\mathrm{g}^T & \mbf{b}_\mathrm{a}^T
    \end{bmatrix}^T,
\end{equation}
where $\pos{W}{S}$ is the position of $\cframe{S}$ in $\cframe{W}$, $\q{W}{S}$ is the Hamiltonian quaternion for orientation of $\cframe{S}$ relative to $\cframe{W}$, $\vel{W}$ is the velocity of $\cframe{S}$ relative to $\cframe{W}$, and $\mbf{b}_\mathrm{g}, \: \mbf{b}_\mathrm{a}$ are gyroscope and accelerometer biases, respectively. Using homogeneous representation (in \textit{italic} font), the coordinate transformation from $\cframe{B}$ to $\cframe{A}$ can be written as $\posh{A}{S} = \T{A}{B} \posh{B}{S}$ with $\T{A}{B} \in \text{SE}(3)$. 
\section{VISUAL-INERTIAL ESTIMATOR}
\subsection{Overview}
We introduce stereo and MVS networks, which predict depths as well as depth uncertainty, into visual-inertial state estimation to probabilistically integrate depths in occupancy mapping and weight occupancy-to-point (map-to-map and frame-to-map) factors in the factor graph optimization. Fig. \ref{fig:block} illustrates our approach: given multiple images and inertial measurements, we optimally fuse depths based on their predicted uncertainties. The fused depth map with its pose is integrated into the currently expanding submap. The most overlapping submaps to the latest submap are found based on its covisible submaps. Map-to-map factors are built from a pair of submaps, while live depths from the stereo network are used to compute the frame-to-map factor. These factors are weighted by predicted uncertainties from the networks. We describe each block in detail in the following sections.

\subsection{Occupancy submapping}
Each submap maintains occupancy log-odds in its coordinate frame where our submapping is based on Supereight2~\cite{SE2}, an octree-based adaptive multi-resolution mapping pipeline. Given the camera pose $\T{M}{C}$ and the depth image $\mbf{D}$, the log-odd of a point in the map frame, $\leftidx{_{M}}{\mathbf{p}}$ is defined as
\begin{equation}
    l\left( \leftidx{_{M}}{\mathbf{p}} \right) = \text{log} \frac{P_{\text{occ}}\left( \leftidx{_{M}}{\mathbf{p}} \:|\: \T{M}{C}, \mbf{D} \right)}{1 - P_{\text{occ}}\left( \leftidx{_{M}}{\mathbf{p}} \:|\: \T{M}{C}, \mbf{D} \right)},
\end{equation}
where $P_\text{occ}$ is the occupancy probability. To account for the uncertainty of depth measurements, we use the piece-wise linear inverse sensor model to approximate the log-odd
\begin{equation}
    l\left( \leftidx{_{M}}{\mathbf{p}} \right) = \begin{cases}
    l_\text{min} & \quad d_r < -3\,\sigma_d, \\
    \frac{|l_\text{min}|}{3\sigma_d}d_r & \quad -3\,\sigma_d \le d_r < \frac{\tau}{2}, \\
    l_\text{max} & \quad \frac{\tau}{2} \le d_r < \tau. \label{eq:l}
    \end{cases}
\end{equation}
Here, $d_r$ stands for the distance between a queried point and the surface measurement along the ray, $l_\text{min}$ is a configurable parameter for the minimum log-odd, $\sigma_d$ is the standard deviation of the depth, $\tau$ stands for surface thickness, which is assumed to be proportional to the depth measurement, and $l_\text{max}$ can be determined from the continuity. To integrate multiple measurements, the mean of log-odd $L(\cdot)$ is recursively updated as
\begin{align}
    L_k\left( \leftidx{_{M}}{\mathbf{p}} \right) &= \frac{L_{k-1}\left( \leftidx{_{M}}{\mathbf{p}} \right) w_{k-1} + l\left( \leftidx{_{M}}{\mathbf{p}} \right)}{w_{k-1}+1}, \nonumber \\
    w_{k} &= \mathrm{min}  \{ w_{k-1}+1, \: w_\text{max} \}, \label{eq:L}
\end{align}
where $w_k$ is the number of observations and saturated in $w_\text{max}$. Supereight2 uses a heuristic uncertainty model to determine $\sigma_d$, such as a linear or a quadratic model. However, this approximation is not valid in general: for instance, textureless walls or object edges could cause highly uncertain depth prediction. In response, we propose to exploit pixel-wise learned uncertainty as an additional output of our deep neural network to integrate noisy depths into occupancy.

\subsection{Factor graph optimization}
Our visual-inertial state estimator is based on OKVIS2~\cite{leutenegger2022okvis2}, which is a visual-inertial SLAM formulated in a factor graph,  and the extended work with LiDAR~\cite{boche2024tightly}. The estimator jointly minimizes the following cost function
\begin{align}
    c(\mathbf{x}) &= \frac{1}{2} \sum_i \sum_{k \in \mathcal{K}} \sum_{j \in \mathcal{J}(i,k)} \rho_\text{c} \left( {\mathbf{e}^{i,j,k}_\mathrm{r}}^T \mathbf{W}_\mathrm{r} \mathbf{e}^{i,j,k}_\mathrm{r} \right) \nonumber \\
    &+ \frac{1}{2} \sum_{k \in \mathcal{P} \cup \mathcal{K} \setminus f} {\mathbf{e}^k_\mathrm{s}}^T \mathbf{W}_\mathrm{s}^k \mathbf{e}^k_\mathrm{s}
    + \frac{1}{2} \sum_{r \in \mathcal{P}} \sum_{c \in \mathcal{C}(r)} {\mathbf{e}^{r,c}_\mathrm{p}}^T \mathbf{W}^{r,c}_\mathrm{p} \mathbf{e}^{r,c}_\mathrm{p} \nonumber \\
    &+ \frac{1}{2} \sum_{k \in \mathcal{P} \cup \mathcal{K}} \rho_\text{t} \left( {\mathbf{e}^{k,o(k)}_{\mathrm{o2p}}}^T \mathbf{W}^{k,o(k)}_{\mathrm{o2p}} \mathbf{e}^{k,o(k)}_{\mathrm{o2p}} \right) \nonumber \\ 
    &+ \frac{1}{2} \sum_{m \in \mathcal{M}} \rho_\text{t} \left( {\mathbf{e}^{m,o(m)}_{\mathrm{o2p}}}^T \mathbf{W}^{m,o(m)}_{\mathrm{o2p}} \mathbf{e}^{m,o(m)}_{\mathrm{o2p}} \right). \label{eq:vi_loss}
\end{align}
In this expression, the reprojection error $\mathbf{e}^{i,j,k}_{\mathrm{r}}$  with the Cauchy robustifier $\rho_\text{c}(\cdot)$ runs over the $i^\text{th}$ camera, the set $\mathcal{K}$ including recent frames and keyframes in the past, and the visible landmark set in those frames $\mathcal{J}(i,k)$. The IMU preintegration error $\mathbf{e}_\mathrm{s}^k$ is formulated starting from poses of $\mathcal{K}$ and the pose graph set $\mathcal{P}$ except from the current frame $f$. The relative pose error $\mathbf{e}^{r,c}_\mathrm{p}$ between the $r^\text{th}$ pose graph frame and the $c^\text{th}$ frame in the set $\mathcal{C}(r)$, which has edges to $r$, is also added. The weighting matrices $\mathbf{W}_\mathrm{r}, \mathbf{W}_\mathrm{s}, \mathbf{W}_\mathrm{p}$ are obtained from the inverse of the reprojection covariance matrix, IMU uncertainty propagation, and the landmark marginalization, respectively. The occupancy-to-point factor $\mathbf{e}_{\mathrm{o2p}}$ with the Tukey loss function $\rho_\text{t}(\cdot)$ encodes the fact that the point should lie on the surface ($L=0$), and it includes $N$ depth measurements as
\begin{align}
    \mathbf{e}^{a,b}_{\mathrm{o2p}} &= 
    \begin{bmatrix}
        e_{\mathrm{o2p}_1} & \cdots & e_{\mathrm{o2p}_N}
    \end{bmatrix}^T, \nonumber \\
    e_{\mathrm{o2p}} &= \frac{L\left( \T{S_b}{W} \T{W}{S_a} \leftidx{_{S_a}}{\mbfh{p}} \right)}{\norm{\nabla L(\leftidx{_{S_b}}{\mathbf{p}})}}.
\end{align}
In this expression, $\mathbf{p}$ is obtained by backprojecting pixels with the network depth and $\nabla L(\cdot)$ stands for the gradient of the log-odd $L$, defined in (\ref{eq:L}).  Depending on which frame is used for the \textit{point}, we have either frame-to-map $\mathbf{e}^{k,o(k)}_{\mathrm{o2p}}$ or map-to-map $\mathbf{e}^{m,o(m)}_{\mathrm{o2p}}$ factors~\cite{boche2024tightly}. Specifically, $\mathbf{e}^{k,o(k)}_{\mathrm{o2p}}$ is formulated by the $k^\text{th}$ frame with its most overlapping submap $o(k)$, while $\mathbf{e}^{m,o(m)}_{\mathrm{o2p}}$ is defined by the $m^\text{th}$ submap with its most overlapping submap $o(m)$. In this paper, the frame-to-map factor constrains the current frame from the stereo network so that the factor can be built before the optimization, while the map-to-map factor aligns submaps using the fused depth that was integrated in the $m^\text{th}$ submap.

The occupancy-to-point weighting matrix with the assumption of linear occupancy near the surface is defined as
\begin{align}
    \mathbf{W}_\mathrm{o2p}^{a,b} &= \text{diag} \left( \begin{bmatrix}
        w_{\mathrm{o2p}_1} & \cdots & w_{\mathrm{o2p}_N}
    \end{bmatrix} \right), \nonumber \\
    w_{\mathrm{o2p}_i} &= \frac{1}{\sigma^2_{\mathrm{map}_i} + \sigma^2_{d_i}}, \:\:\: \sigma_{\mathrm{map}_i} = \frac{l_\text{min}}{\norm{3\nabla L(\leftidx{_{S_b}}{\mathbf{p}})}},
\end{align}
where the total uncertainty is the sum of the occupancy map uncertainty $\sigma_{\mathrm{map}}$ and the network depth uncertainty $\sigma_d$. $l_\text{min}$ stands for the minimum log-odd value as defined in (\ref{eq:l}). In contrast to highly precise LiDAR measurements, the depth uncertainty, which often includes noisy observations, should be assigned pixel-wise to properly weight its contribution to the optimization loss. In the following section, we describe how the uncertainty is obtained from networks.

\subsection{Uncertainty-aware depth fusion}
It is pivotal to obtain a reliable dense depth as well as the associated uncertainty for our downstream tasks. Our key idea is to fuse static and motion stereo, which are complementary to each other --- static stereo provides a reliable small baseline even when stationary, while motion stereo potentially brings a large baseline, depending on the camera motion. Our method does not depend on a specific network, but we found that Unimatch~\cite{xu2023unifying} and the MVS network~\cite{xin2023simplemapping} work well in real-world scenes. However, the previous networks only predict disparity or depth without any uncertainties. Therefore, we augment the base architecture with an uncertainty decoder and adopt the Laplacian loss function for (aleatoric) uncertainty learning. Specifically, our stereo network loss function is
\begin{equation}
    \mathcal{L}_{\text{st}}(\mbf{\theta}) = \mathcal{L}_{u}(\mbf{\theta}) + \mathcal{L}_{\nabla u_x}(\mbf{\theta}) + \mathcal{L}_{\nabla u_y}(\mbf{\theta}) \label{eq:stereo_loss},
\end{equation}
where $\mbf{\theta}$ is the network weights, and $u$ stands for the disparity. The disparity loss $\mathcal{L}_u$, modeled in the Laplacian distribution as in~\cite{kendall2017uncertainties, poggi2020uncertainty}, is defined as
\begin{equation}
    \mathcal{L}_{u}(\mbf{\theta}) = \sum_{i\in\mathcal{T}} \frac{\abs{u_i - u_{\text{gt}_i}}}{\sigma_{u_i}} + \log{\sigma_{u_i}},
\end{equation}
where $\mathcal{T}$ is a training set including pairs of stereo images and the ground-truth disparity. We additionally add the gradient loss $\mathcal{L}_{\nabla u}$ for sharper uncertainty output, which is analogously defined as $\mathcal{L}_u$. The gradient uncertainty along the horizontal and vertical directions is derived from the disparity uncertainty as
\begin{align}
    \sigma_{\nabla u_x} &= \sqrt{\sigma^2_{u}(x+1,y) + \sigma^2_{u}(x-1,y)}, \nonumber \\
    \sigma_{\nabla u_y} &= \sqrt{\sigma^2_{u}(x,y+1) + \sigma^2_{u}(x,y-1)}.
\end{align}
We propagate the uncertainty from the disparity to the depth with linearization,
\begin{align}
    \hat{d}_{\text{st}} = \frac{f_c b}{u}, \quad \sigma_{\text{st}} = \frac{f_c b}{u^2} \sigma_u,
\end{align}
where $f_c$ is the rectified focal length, $b$ is the stereo baseline.

Likewise, we modify the loss function of the MVS network~\cite{xin2023simplemapping}
\begin{equation}
    \mathcal{L}_{\text{mvs}}(\mbf{\phi}) = \sum_{i\in\mathcal{T}} \frac{\abs{\log{d_i}-\log{d_{\text{gt}_i}}}}{\sigma_{l_i}} + \log{\sigma_{l_i}},
\end{equation}
where the network learns log-depth uncertainty. We transform the log-depth to a depth with a linearized model,
\begin{align}
    \hat{d}_{\text{mvs}} = \exp{\left(\log{d_i}\right)}, \quad \sigma_{\text{mvs}} = \hat{d}_{\text{mvs}} \sigma_l. \label{eq:mvs_loss}
\end{align}

\begin{figure}
\centerline{\includegraphics[width=.95\linewidth]{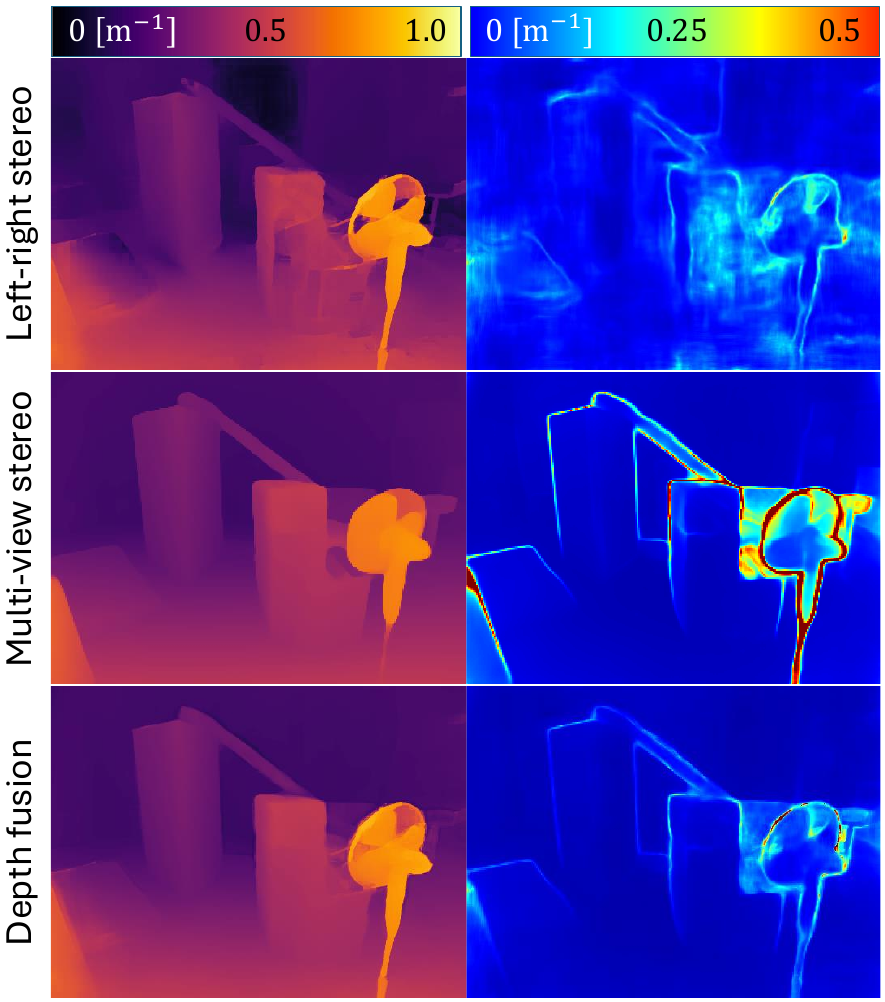}}
\caption{Predicted inverse (only for visualization purpose) depth and its corresponding standard deviation of (top) stereo network with the $11\,\text{cm}$ baseline, (middle) MVS network with the $50\,\text{cm}$ maximum baseline among $8\,$views, and (bottom) depth fusion in the EuRoC dataset.} \label{fig:euroc_images}
\end{figure}

Given the pixel-wise estimates from the networks $(\hat{d}_\text{st}, \sigma_\text{st})$, $(\hat{d}_\text{mvs}, \sigma_\text{mvs})$ and with the assumption that two estimates are independent, we can optimally fuse two depth estimates~\cite{carlson1990federated},
\begin{align}
    \hat{d}_{\text{fuse}} &= \sigma^2_\text{fuse} \left( \sigma^{-2}_\text{st} \hat{d}_{\text{st}} + \sigma^{-2}_\text{mvs} \hat{d}_{\text{mvs}}\right), \nonumber \\
    \sigma^2_{\text{fuse}} &= \left( \sigma^{-2}_{\text{st}} + \sigma^{-2}_{\text{mvs}}\right)^{-1}.
\end{align}
Fig. \ref{fig:euroc_images} shows an example where the fan stand has more detailed depth in the stereo network, while farther objects are sharper in the MVS network. Fused depth naturally maintains the optimal depth based on uncertainty-aware fusion.

\subsection{Submap management}
It is crucial to maintain uniformly distributed submaps with sufficient, yet not redundant overlaps between consecutive submaps for the occupancy-to-point factor. To achieve this, we keep tracking the overlap volume between the current depth frustum and the latest submap. If this falls below a certain threshold $\epsilon_\text{vol}$, a new submap is generated. We approximate this by sampling points in the frustum and test whether points have been observed in the previous submap.

We build the occupancy-to-point factors based on the the oldest covisible submap for geometric loop closure. We record covisible submaps to the latest submap in terms of co-observed sparse landmarks. Then, the most overlapping submap is declared as the oldest one among covisible submaps with at least $M_\text{min}$ points. The submap poses are updated accordingly after optimizing the factor graph (\ref{eq:vi_loss}), in order to maintain a globally consistent map.
\section{EXPERIMENTAL RESULTS}
We evaluate our method on two public datasets, EuRoC~\cite{burri2016euroc} and Hilti-Oxford~\cite{hilti2022} in terms of localization and mapping accuracy. The estimated trajectory is aligned in $\text{SE}(3)$ to the ground-truth before evaluation. We report localization accuracy as absolute trajectory error (ATE) RMSE in EuRoC or the predefined score where the position error $[1,10]\text{cm}$ is scaled to $[0,100]$ points in the Hilti-Oxford dataset. We report causal and non-causal errors where the former only takes measurements up to the current time, while the latter is the final loop-closed trajectory~\cite{leutenegger2022okvis2}. To evaluate mapping accuracy, submap meshes are reconstructed using marching cubes~\cite{lorensen1998marching} and combined using submap poses. Estimated and ground-truth vertices are downsampled with a voxel size of $1\text{cm}^3$, then point-to-plane ICP aligns both sets of vertices. We report accuracy as an average distance from all estimated vertices to the ground-truth without threshold in meter and completeness as a fraction of ground-truth vertices that are within $0.2\text{m}$ to the estimated vertices.

\begin{table}
\caption{Absolute Trajectory Error [m] in the EuRoC dataset} \label{tab:euroc_ate}
\centering
\renewcommand{\arraystretch}{1.15} % Default value: 1
\begin{threeparttable}[b]
\begin{tabular}{c|c c c c|c c c}
    & \rotatebox[origin=c]{90}{VINS-Fusion\tnote{1}~\cite{qin2019general}} & \rotatebox[origin=c]{90}{OKVIS2\tnote{2}~\cite{leutenegger2022okvis2}} & \rotatebox[origin=c]{90}{DVI-SLAM\tnote{3}~\cite{peng2023dvi}} & \rotatebox[origin=c]{90}{Ours} & \rotatebox[origin=c]{90}{ORB-SLAM3\tnote{3}~\cite{campos2021orb}} & \rotatebox[origin=c]{90}{OKVIS2\tnote{2}~\cite{leutenegger2022okvis2}} & \rotatebox[origin=c]{90}{Ours} \\ \\[-1em]
    
    \hline
    & \multicolumn{4}{c|}{\textit{Causal}} & \multicolumn{3}{c}{\textit{Non-causal}} \\

    \hline
    \texttt{MH01} & 0.166 & 0.034 & 0.042 & {0.034} & 0.036 & \underline{0.021} & \textbf{0.019} \\
    \texttt{MH02} & 0.152 & {0.029} & 0.046 & {0.031} & 0.033 & \underline{0.020} & \textbf{0.019} \\
    \texttt{MH03} & 0.125 & {0.039} & 0.081 & 0.037 & 0.035 & \underline{0.032} & \textbf{0.029} \\
    \texttt{MH04} & 0.280 & 0.065 & 0.072 & {0.072} & \textbf{0.051} & \underline{0.059} & 0.071 \\
    \texttt{MH05} & 0.284 & 0.087 & {0.069} & 0.081 & 0.082 & \textbf{0.061} & \underline{0.066} \\

    \hline
    \texttt{V101} & 0.076 & \underline{0.037} & 0.059 & \textbf{0.035} & 0.038 & \textbf{0.035} & \textbf{0.035} \\
    \texttt{V102} & 0.069 & {0.031} & 0.034 & \underline{0.028} & \textbf{0.014} & \textbf{0.014} & \textbf{0.014} \\
    \texttt{V103} & 0.114 & 0.035 & {0.028} & 0.029 & 0.024 & \textbf{0.020} & \underline{0.021} \\
    
    \hline
    \texttt{V201} & 0.066 & {0.039} & 0.040 & 0.033 & 0.032 & \textbf{0.023} & \underline{0.024} \\
    \texttt{V202} & 0.091 & {0.030} & 0.039 & 0.028 & \textbf{0.014} & 0.019 & \underline{0.016} \\
    \texttt{V203} & 0.096 & 0.041 & 0.055 & {0.041} & \underline{0.024} & 0.027 & \textbf{0.022} \\
    
    \hline
    Avg & 0.138 & 0.043 & 0.051 & {0.041} & \underline{0.035} & \textbf{0.030} & \textbf{0.030} \\

    \hline
\end{tabular}
\begin{tablenotes}
   \item [1] Results taken from~\cite{campos2021orb}.
   \item [2] Results obtained ourselves.
   \item [3] Results taken from the paper.
 \end{tablenotes}
\end{threeparttable}
\end{table}

\begin{table}
\caption{Mesh accuracy [m] without threshold and completeness [\%] with 0.2m threshold in the EuRoC dataset} \label{tab:euroc_mesh}
\centering
\renewcommand{\arraystretch}{1.15} % Default value: 1
\begin{threeparttable}[b]
\begin{tabular}{c|c c c|c c c}
     & \rotatebox[origin=c]{90}{Simplemapping\tnote{1}~\cite{xin2023simplemapping}} & \rotatebox[origin=c]{90}{Ours-Mono} & \rotatebox[origin=c]{90}{Ours} & \rotatebox[origin=c]{90}{Simplemapping\tnote{1}~\cite{xin2023simplemapping}} & \rotatebox[origin=c]{90}{Ours-Mono} & \rotatebox[origin=c]{90}{Ours} \\ \\[-1em]
    
    \hline
    & \multicolumn{3}{c|}{\textit{Accuracy}} & \multicolumn{3}{c}{\textit{Completeness}} \\

    \hline
    \texttt{V101} & 0.176 & \underline{0.076} & \textbf{0.042} & 40.58 & \underline{41.74} & \textbf{44.28} \\
    \texttt{V102} & 0.123 & \underline{0.073} & \textbf{0.041} & \underline{55.07} & 54.96 & \textbf{58.38} \\
    \texttt{V103} & 0.161 & \underline{0.093} & \textbf{0.048} & 47.74 & \underline{55.92} & \textbf{68.48} \\
    
    \hline
    \texttt{V201} & 0.155 & \underline{0.090} & \textbf{0.072} & \underline{41.93} & 39.62 & \textbf{43.64} \\
    \texttt{V202} & 0.156 & \underline{0.073} & \textbf{0.069} & \underline{56.37} & 54.79 & \textbf{57.66} \\
    \texttt{V203} & {0.090} & \underline{0.089} & \textbf{0.072} & \textbf{62.58} & 58.42 & \underline{59.96} \\
    
    \hline
    Avg & 0.144 & \underline{0.082} & \textbf{0.057} & 50.71 & \underline{50.91} & \textbf{55.40} \\

    \hline
\end{tabular}
\begin{tablenotes}
   \item [1] Results obtained ourselves with the same metric.
 \end{tablenotes}
\end{threeparttable}
\end{table}

\begin{figure}
\centerline{\includegraphics[width=.98\linewidth]{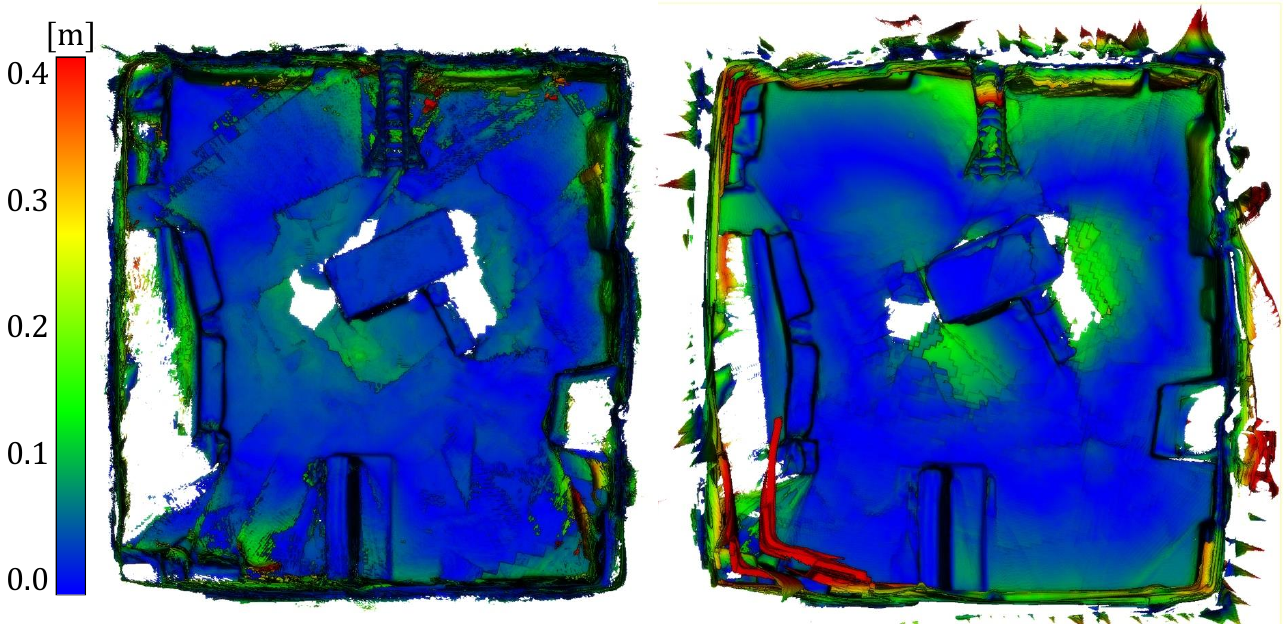}}
\caption{Reconstructed mesh with mesh-to-point error encoding in \texttt{V102} sequence. (Left) Ours and (Right) Simplemapping~\cite{xin2023simplemapping}.} \label{fig:euroc_mesh}
\end{figure}

\subsection{Implementation Details}
We fine-tune the stereo network~\cite{xu2023unifying} starting from pre-trained weights on the TartanAir dataset~\cite{wang2020tartanair} without the post-refinement for fast inference. Then, we add a decoder adopted from MonoDepth2~\cite{godard2019digging} after the correlation volume for uncertainty learning while freezing the other weights in the loss (\ref{eq:stereo_loss}). Likewise, we add an uncertainty decoder to the MVS network~\cite{xin2023simplemapping}, which has the same architecture as the depth decoder of our base network. Then, we train uncertainty starting from the pre-trained weights using the loss (\ref{eq:mvs_loss}) while freezing the rest of weights in the ScanNet dataset~\cite{dai2017scannet}. After training, we calibrate the uncertainty prediction by multiplying by a constant gain to ensure that the averaged Mahalanobis distance equals the unit distance in the validation set.

Incoming images are downsampled to a 512$\times$384 resolution in $10\,\text{Hz}$ for the networks. We input 8 images with up-to-date poses after optimizing the factor graph with a varying baseline of $3\,\text{cm}$ or $10\,\text{cm}$ and a sparse landmark depth map to our MVS network. The voxel resolution is set to $2.5\,\text{cm}$ at each submap. We set $l_\text{min}=-5.015$, $\epsilon_\text{vol}=0.2$, $M_\text{min}=100$ and take maximum of 200 and 1000 points for the frame-to-map and map-to-map factors, respectively. We downsample the points with the lowest uncertainty if the number of points is above the threshold. We use Ceres solver~\cite{Agarwal_Ceres_Solver_2022} for the optimization and use the same parameters per dataset. All the experiments were run on a PC with an Intel i7-13700 CPU and an Nvidia RTX3080 GPU.

% how training is done; how the network was calibrated; how mvs works with OKVIS2; how rectified; Disclose all parameters presented during method; how many points are using for the submap alignment.

\subsection{The EuRoC Dataset}
This dataset provides stereo images and IMU measurements recorded by a drone, the ground-truth trajectory and point clouds of the Vicon room in mm-level accuracy~\cite{burri2016euroc}. We use the stereo pair for our stereo network and left images for our MVS network. Table \ref{tab:euroc_ate} summarizes ATE where all competitors also use a stereo and inertial configuration. Our method improves the baseline, OKVIS2~\cite{leutenegger2022okvis2} by adding occupancy-to-point factors in causal evaluation. We obtained the same accuracy on average when compared to the baseline in non-causal evaluation. This indicates that the depth fusion could not achieve $3\,\text{cm}$-level accuracy, which is already quite accurate. On the other hand, our approach outperforms state-of-the-art methods with nontrivial margins. Since our method tightly couples localization and volumetric mapping, we also evaluate the mapping accuracy in Table \ref{tab:euroc_mesh}. For fair comparison to Simplemapping~\cite{xin2023simplemapping} in a monocular-inertial set up, we implement \textit{Ours-Mono} where only left images are used without the stereo network. The proposed method achieves the highest accuracy and completeness, while \textit{Ours-Mono} also outperforms the competitor. Fig. \ref{fig:euroc_mesh} compares the mapping accuracy where \textit{Ours} improves mesh accuracy especially along the edge of the room.

\begin{table}
\centering
\begin{threeparttable}[b]
\caption{Localization scores in the Hilti-Oxford dataset} \label{tab:hilti}
\centering
\renewcommand{\arraystretch}{1.15} % Default value: 1
\begin{tabular}{c|c c|c c c c}
    & \rotatebox[origin=c]{90}{OKVIS2\tnote{1}~\cite{leutenegger2022okvis2}} & 
    \rotatebox[origin=c]{90}{Ours} &
    \rotatebox[origin=c]{90}{BAMF-SLAM\tnote{2}~\cite{zhang2023bamf}} &
    \rotatebox[origin=c]{90}{XR Penguin\tnote{2}~\cite{wang2024mavis}} & 
    \rotatebox[origin=c]{90}{OKVIS2\tnote{1}~\cite{leutenegger2022okvis2}} & \rotatebox[origin=c]{90}{Ours} \\ \\[-1em]
    
    \hline
    & \multicolumn{2}{c|}{\textit{Causal}} & \multicolumn{4}{c}{\textit{Non-causal}} \\

    \hline
    \texttt{exp01} & 8.46 & 4.62 & 15.38 & 40.00 & \underline{45.38} & \textbf{55.38} \\
    \texttt{exp02} & 4.09 & 0.45 & 4.55 & \underline{16.82} & 15.91 & \textbf{24.09}  \\
    \texttt{exp03} & 0.00 & 0.00 & \underline{8.24} & \textbf{16.47} & 0.00 & 0.00 \\
    \texttt{exp07} & 6.67 & \underline{15.00} & 1.67 & {11.67} & \underline{15.00} & \textbf{21.67} \\
    \texttt{exp09} & 5.00 & 2.50 & 1.25 & \underline{15.00} & 10.62 & \textbf{16.88} \\
    \texttt{exp11} & 4.00 & 2.00 & 2.00 & 20.00 & \underline{36.00} & \textbf{38.00} \\
    \texttt{exp15} & 14.44 & 20.00 & 7.78 & 24.67 & \textbf{47.78} & \underline{43.33} \\
    \texttt{exp21} & 0.00 & 0.00 & 0.00 & \textbf{4.00} & 0.00 & 0.00 \\
    
    \hline
    Total & 42.66 & 44.57 & 40.86 & 150.62 & \underline{170.70} & \textbf{199.35} \\

    \hline
\end{tabular}
\begin{tablenotes}
   \item [1] Results obtained ourselves with online extrinsics calibration.
   \item [2] Results taken from the leaderboard~\cite{hilti_leaderboard}.
 \end{tablenotes}
\end{threeparttable}
\end{table}

\subsection{The Hilti-Oxford dataset}
This dataset provides images from 5 cameras and IMU measurements captured by a handheld device, sparse ground-truth positions, and mm-accurate dense point clouds~\cite{hilti2022}. We use all 5 cameras for the visual-inertial estimator, 2 front cameras for the stereo network, and the front left camera for the MVS network. We calibrate camera-IMU extrinsic parameters online. Table \ref{tab:hilti} reports the localization score in the Challenge sequences where our non-causal implementation also includes the final bundle adjustment. At time of writing, our method is ranked first place on the leaderboard among the published methods. In terms of mean position error, our causal score corresponds to $13.3\,\text{cm}$ and the non-causal score to $6.1\,\text{cm}$ in sequences from which the score was obtained. Fig. \ref{fig:exp04_slam} visualizes the estimated trajectory in colored submap meshes in \texttt{exp04} where the ground-truth point clouds are available --- our method gives $9.4\text{cm}$ accuracy and $62.2\%$ completeness.

\subsection{Ablation Study}
This study shows the effectiveness of our contributions. Table \ref{tab:euroc_albation} summarizes averages over all sequences in the EuRoC dataset. The localization accuracy has been improved with uncertainty-awareness. Furthermore, we improve mapping accuracy $10\%$ by having network-predicted uncertainty over the heuristic quadratic uncertainty~\cite{SE2} and $29\%$ by the depth fusion. Fig. \ref{fig:exp04_slam} qualitatively shows less noisy and more complete meshes of our depth fusion when compared to the stereo network with a heuristic uncertainty model.

\subsection{Timings and memory}
We measure computation time in \texttt{MH01} of the EuRoC dataset as shown in Table \ref{tab:euroc_timing}. The real-time estimator of our method with loop-closure optimizations running in the background can run in $29\,\text{Hz}$, while OKVIS2 as our baseline can run in $41\,\text{Hz}$. Our mapping can run at $13\,\text{Hz}$ given the bottleneck of the MVS network. Also, MVS and stereo networks only take 3.51GB of GPU memory during inference.

\begin{table}
\caption{Ablation study: ATE in causal (C), non-causal (NC), accuracy (Acc.), and completeness (Comp.) in EuRoC dataset} \label{tab:euroc_albation}
\centering
\setlength{\tabcolsep}{4pt} % Default value: 6pt
\renewcommand{\arraystretch}{1.15} % Default value: 1
\begin{tabular}{l c c c| c c c c}
    \hline
    & \multirow{2}{*}{S} & \multirow{2}{*}{U} & \multirow{2}{*}{F} & ATE-C & ATE-NC & {Acc.} & {Comp.} \\[-2pt]
    & & & & [m] & [m] & [m] & [\%]\\

    \hline
    Baseline~\cite{leutenegger2022okvis2} & & & & 0.043 & \textbf{0.030} & - & - \\
    + Stereo network & \checkmark & & & 0.045 & 0.032 & 0.080 & 50.38 \\
    + Uncertainty & \checkmark & \checkmark & & 0.043 & 0.031 & 0.072 & 50.31 \\
    + Depth fusion & \checkmark & \checkmark & \checkmark & \textbf{0.041} & \textbf{0.030} & \textbf{0.057} & \textbf{55.40} \\

    \hline
\end{tabular}
\end{table}

\begin{table}
\caption{Timing in mean$\pm\sigma\,$[ms] measured in the EuRoC dataset} \label{tab:euroc_timing}
\centering
\setlength{\tabcolsep}{4pt} % Default value: 6pt
\renewcommand{\arraystretch}{1.15} % Default value: 1
\begin{tabular}{c| c c c c}
    \hline
    & \multirow{2}{*}{VI optimization} & Occupancy & Stereo & MVS \\[-3pt]
    & & update & network & network \\
    \hline
    OKVIS2~\cite{leutenegger2022okvis2} & 24.2$\,\pm\,$5.3 & - & - &  - \\
    Ours & 34.3$\,\pm\,$6.6 & 11.6$\,\pm\,$3.6 & 56.3$\,\pm\,$24.1 & 75.9$\,\pm\,$9.7 \\

    \hline
\end{tabular}
\end{table}
\section{CONCLUSION}
We have proposed visual-inertial SLAM that tightly couples localization and volumetric occupancy mapping through uncertainty-aware occupancy-to-point factors. To further improve the mapping accuracy, we fine-tune stereo and MVS networks with uncertainty learning. Then, we optimally fuse depths from these networks based on their predicted uncertainty, which is employed in the optimization and depth integration in submapping. Through evaluation in benchmark datasets, our method outperforms the state-of-the-art regarding accuracy both in localization and mapping, while providing dense occupancy in real-time ($13\text{Hz}$). For future work, we aim to consider dynamic objects, incorporate epistemic depth uncertainty, and close the loop with robot navigation and control accounting for the state uncertainty.

% Limitation: only consider aleatoric uncertainty, MVS network is not generalizable in open-sky region; decouple MVS network in real-time

\clearpage
\bibliographystyle{IEEEtran}
\bibliography{bib/IEEEabrv, bib/myIEEE.bib}

\end{document}